\pdfoutput=1

\documentclass[11pt]{article}

\usepackage[preprint]{coling}

\usepackage{times}
\usepackage{latexsym}
\usepackage{ccicons}

\usepackage[T1]{fontenc}

\usepackage[utf8]{inputenc}

\usepackage{microtype}

\usepackage{inconsolata}

\usepackage{epigraph}
\usepackage{graphicx}
\usepackage{tabularx}
\usepackage{graphics}
\usepackage{enumitem}
\usepackage{multirow}
\usepackage{multicol}
\usepackage{float}
\usepackage{hyperref}
\usepackage{amsmath}
\usepackage{amssymb}
\usepackage{booktabs}
\usepackage{xtab}
\usepackage{svg}

\usepackage{varwidth}

\title{Transforming Scholarly Landscapes: \\ Influence of Large Language Models on Academic Fields \\ beyond Computer Science}

\author{
Aniket Pramanick\textsuperscript{1}, Yufang Hou\textsuperscript{2}, Saif M. Mohammad\textsuperscript{3}, Iryna Gurevych\textsuperscript{1} \\
\textsuperscript{1}Ubiquitous Knowledge Processing Lab (UKP Lab) \\
Department of Computer Science and Hessian Center for AI (hessian.AI) \protect\\
\textsuperscript{2}IBM Research Europe, Ireland \\
\textsuperscript{3}National Research Council Canada \\
\texttt{\url{www.ukp.tu-darmstadt.de}}
}

\begin{document}
\maketitle

\begin{abstract}

Large Language Models (LLMs) have ushered in a transformative era in Natural Language Processing (NLP), reshaping research 
and extending NLP's influence to other fields of study. 
However, there is little to no work examining
the degree to which LLMs influence other research fields.
This work {\it empirically and systematically examines the influence and use of LLMs in fields beyond NLP.}
We curate $106$ LLMs and analyze $\sim$$148k$ papers citing LLMs to quantify their influence and reveal trends in their usage patterns.
Our analysis reveals not only the increasing prevalence of LLMs in non-CS fields but also the disparities in their usage, with some fields utilizing them more frequently than others since 2018, notably Linguistics and Engineering together accounting for $\sim$$45\%$ of LLM citations. 
Our findings further indicate that most of these fields predominantly employ task-agnostic LLMs, proficient in zero or few-shot learning without requiring further fine-tuning, to address their domain-specific problems. 
This study sheds light on the cross-disciplinary impact of NLP through LLMs, providing a better understanding of the opportunities and challenges.\footnote{code and data available at \url{https://github.com/UKPLab/arxiv-2024-llm-trends}}


\end{abstract}

\section{Introduction}

\epigraph{``What you do makes a difference, and you have to decide what kind of difference you want to make.''}{- Jane Goodall}

Modern science heavily relies on citing past research to build upon past ideas, situate the proposed work, reject old hypotheses, etc., thereby facilitating the dissemination of good ideas~\cite{jurgens-etal-2018-measuring}. Some ideas are limited in the scope of their influence (being cited narrowly within a specific subfield). At the same time, others may be broadly applicable and influence not just a whole field of study (such as Computer Science or CS) but also many outside fields. Since the number of published papers is too large for manual review~\citep{fortunato2018science, bornmann2021growth}, there is growing work in automatically and quantitatively tracking the influence of ideas within and across fields~\citep{wahle-etal-2023-cite, wahle2024citation}.

Arguably, one of the most transformative ideas over the past decade is that of LLMs~\citep{li-etal-2023-defining}. Originally proposed within NLP, LLMs have revolutionized almost all research areas within NLP itself~\citep{Radford2019LanguageMA, henderson2023foundation}. Moreover, their utilization is not confined to NLP; other fields are leveraging LLMs as well~\citep{wang2022language, wang2023code4struct}. While it is well recognized that these models are being adopted outside of NLP, the full extent and nature of their usage remains unclear. Analyzing the widespread adoption and utilization of LLMs across various fields provides essential insights for promoting responsible AI practices. Although recent studies have begun to explore the influence of LLMs within CS~\citep{movva2023large, zhu2024we}, their impact on other disciplines beyond CS is still largely unknown. While it is evident that non-CS fields are paying attention to LLMs~\citep{zhao2023survey}, precise details about which fields are leveraging them and the specific purposes of their use are still to be understood.  

The modes of influence are multifaceted and intricate in nature, which makes it challenging to empirically determine the degree of influence. 
In this work, we 
focus on a specific dimension of influence: the scientific impact that one field has on another~\citep{singh-etal-2023-forgotten}. 
One notable marker of this inter-field influence is citation~\citep{zhu2015measuring}. Therefore, we propose that the degree to which a source field cites the works of a target field can serve as a rough indicator of their mutual influence~\citep{siddharthan2007whose}. Although citation patterns are susceptible to 
biases, we can glean meaningful insights 
at an aggregate level~\citep{mohammad2020examining, wahle-etal-2023-citation-field}. 

While no universally accepted definition of LLMs exists~\citep{Radford2019LanguageMA}, in this work, we take LLMs as foundational models~\citep{henderson2023foundation} built on the transformer architecture, pretrained on massive textual datasets with over 100M parameters.
We carefully curated a dataset of $106$ well-cited LLM papers up to February 2024 including BERT \cite{devlin-etal-2019-bert}, T5 \cite{raffel2020exploring}, GPT-3 \cite{brown2020language}, PaLM \cite{chowdhery2022palm}, ChatGPT \cite{gpt35turbo}, and LLaMA \cite{touvron2023llama}. 
We then {\it quantitatively} investigate:\\[-18pt] 

\begin{enumerate}[label={(\Alph*)}]
\item  \emph{Which non-CS fields are impacted by LLMs? And to what degree?} (\S~\ref{subsec:impact_and_influence})\\[-20pt]

\item \emph{How do the usage patterns of LLMs evolve over time within these non-CS fields?} (\S~\ref{subsec:utilization_patterns}) \\[-20pt]
\end{enumerate}

\noindent We complement the above discussions with an additional {\it qualitative} analysis exploring:\\[-20pt]
\begin{enumerate}[label={(\Alph*)}, resume]
\item \emph{In what contexts are LLMs applied within these non-CS fields?} (\S~\ref{subsec:ethical_awareness})
\end{enumerate}
\noindent In addressing the questions above, we utilize the Semantic Scholar data~\citep{lo-wang-2020-s2orc} to construct our dataset comprising $\sim$$148k$ papers from 22 fields outside of CS, published between 2018 and February 2024, that cite LLMs. This dataset includes structured full texts extracted from the PDFs of the papers as well as their metadata, including the fields of study, year of publication, venue of publication, and author information. Additionally, we include similar data from the papers associated with $106$ LLMs, along with $\sim$$273k$ LLM citations from the aforementioned papers outside of CS.

While the quantitative analyses (\textbf{A} and \textbf{B}) provide insights into the adoption of LLMs across diverse fields,
they do not capture the nuances of how LLMs are being utilized within these fields. To gain deeper insights for \textbf{C}, we delve beyond metadata, exploring the content of the papers citing LLMs. Our qualitative examination of papers' contents not only illuminates diverse applications of LLMs outside of CS but also guides towards potential improvements and research directions for better LLM usability in various non-CS fields. Finally, we approximately assess and discuss the extent to which these papers, especially in fields like Biology, Medicine, Psychology, Law, etc., discuss ethical concerns associated with LLMs, such as hallucinations~\citep{shi2023large} or non-reproducible outputs~\citep{de2023evaluation}, using a smaller human-annotated dataset.
In summary, our findings suggest that 1) linguistics, engineering, and medicine are the top three fields citing LLMs, and fields more closely aligned with CS tend to more readily adopt LLM technology compared to other domains (\S~\ref{subsec:impact_and_influence});
2) BERT continues to be the preferred LLM among non-CS fields, even in terms of average yearly citations. Moreover, many of these fields frequently cite task-agnostic models like GPT-3~\citep{brown2020language} or LLaMA~\cite{touvron2023llama}, which excel in few-shot settings without needing additional fine-tuning.
(\S~\ref{subsec:utilization_patterns}); and 3) non-CS fields mostly use LLMs to solve their domain-specific problems (\S~\ref{subsec:ethical_awareness}); 
Our study is the first attempt to empirically and systematically investigate the impact and usage of LLMs in fields outside CS.\\[-10pt] 
\newline \noindent{\bf Study Goal:} The aim of our study is not to engage in debates regarding the advantages or disadvantages of increased LLM usage across diverse fields, nor do we seek to discuss whether and how non-CS fields effectively implement ethical review processes. Instead, our primary objective is to quantify and delineate the utilization of LLMs in fields beyond CS.\\[3pt] 
\noindent \textbf{Motivation and Audience:} NLP researchers play a pivotal role in facilitating interdisciplinary exchange~\citep{hovy-spruit-2016-social}. 
Raising awareness about how NLP technologies are currently being used by other fields is the first step in helping them effectively reach other communities in the future. 
Therefore, {\it the primary audience of this work is NLP researchers}. The insights gleaned here can 
help them craft more effective interdisciplinary projects, foresee potential challenges, and effectively relay these insights and risks to colleagues outside of CS. Moreover, this work could potentially serve as a valuable resource for 
many early NLP researchers, broadening their understanding of NLP's intersection with other domains. Further, we anticipate that this paper will not only benefit the *CL community but also resonate with other research communities, thereby enriching the broader landscape of NLP research.

\section{Related Work}

\paragraph{Scientific Trends Analysis.}
Early work by \citet{hall-etal-2008-studying} served as a catalyst for exploring scientific trends in NLP. Research in scientific trends analysis primarily spans three dimensions, from citation patterns to metadata and content analysis. A vast amount of research has been done into the study of citation patterns and the amalgamation of topological measures within citation networks to assess research trends~\citep{small_et_al, shibata_et_al, boyack_et_al}. Complementary to this, \citet{prabhakaran-etal-2016-predicting} did a content analysis, employing rhetorical framing to elucidate trend patterns. \citet{Grudin_2009} and \citet{Liu2015ExploringTL} take a metadata-driven approach, employing prevalence correlation to examine the interplay between publication topics and research grants. \citet{pramanick-etal-2023-diachronic} apply causal techniques to explore how tasks, methods, datasets, and metrics interact within the context of the evolution of NLP research. \citet{koch2021reduced} explore dataset usage patterns across research communities and observe distinct trends in dataset creation within NLP communities, shedding light on the nuanced dynamics of research resource utilization. 
While the majority of the previous research in Computer Science and NLP examines research trends within Computer Science and NLP, respectively, in this work, we analyze both citations and content to investigate the influence of LLMs (CS and NLP technology) on fields outside of Computer Science. 

\paragraph{NLP Scientometrics.} In parallel, NLP scientometrics has witnessed significant advancements in recent years as researchers strive to understand the landscape of NLP research and its evolution \citep{mingers2015review, chen2019visualizing}. Prior studies have examined the engagement of NLP research with other disciplines, shaping the research trajectory~\citep{hansson1999interdisciplinarity, leydesdorff2019interdisciplinarity}. While various studies have observed a rising trend in interdisciplinary interactions occurring both across fields and within subfields~\citep{van2015interdisciplinary, truc2020interdisciplinarity}, this growth primarily manifests in connections between closely related domains,  with limited increases in associations between traditionally distant research areas, such as materials sciences and clinical medicine~\citep{porter2009science}.
We add to this line of work by investigating how research fields that were traditionally distant from NLP are now incorporating LLM techniques to address their challenges. 


\paragraph{LLM Abilities Assessment.}
LLMs are a category of transformer-based language models distinguished by their size, typically consisting of  hundreds of millions or even more  parameters. 
They are trained 
on vast and diverse data~\citep{shanahan2022talking}. These models exhibit remarkable proficiency in understanding and generating natural language, enabling them to excel in complex language-related tasks~\citep{sanh2021multitask, wei2021finetuned}. However,
LLMs come with challenges, including the generation of plausible yet factually incorrect text~\citep{ganguli2022red}. Moreover, they can be manipulated to produce harmful content, raising ethical concerns about their misuse~\citep{rae2021scaling, du2022glam}. 
Our study does not assess LLM capabilities or risks but focuses on how non-CS fields use LLMs, and we examine ethical concerns related to LLMs based on the content of the papers within these fields. 

\begin{figure*}
   \centering
   \scalebox{0.80}{
    \includegraphics[width=1.0\textwidth]{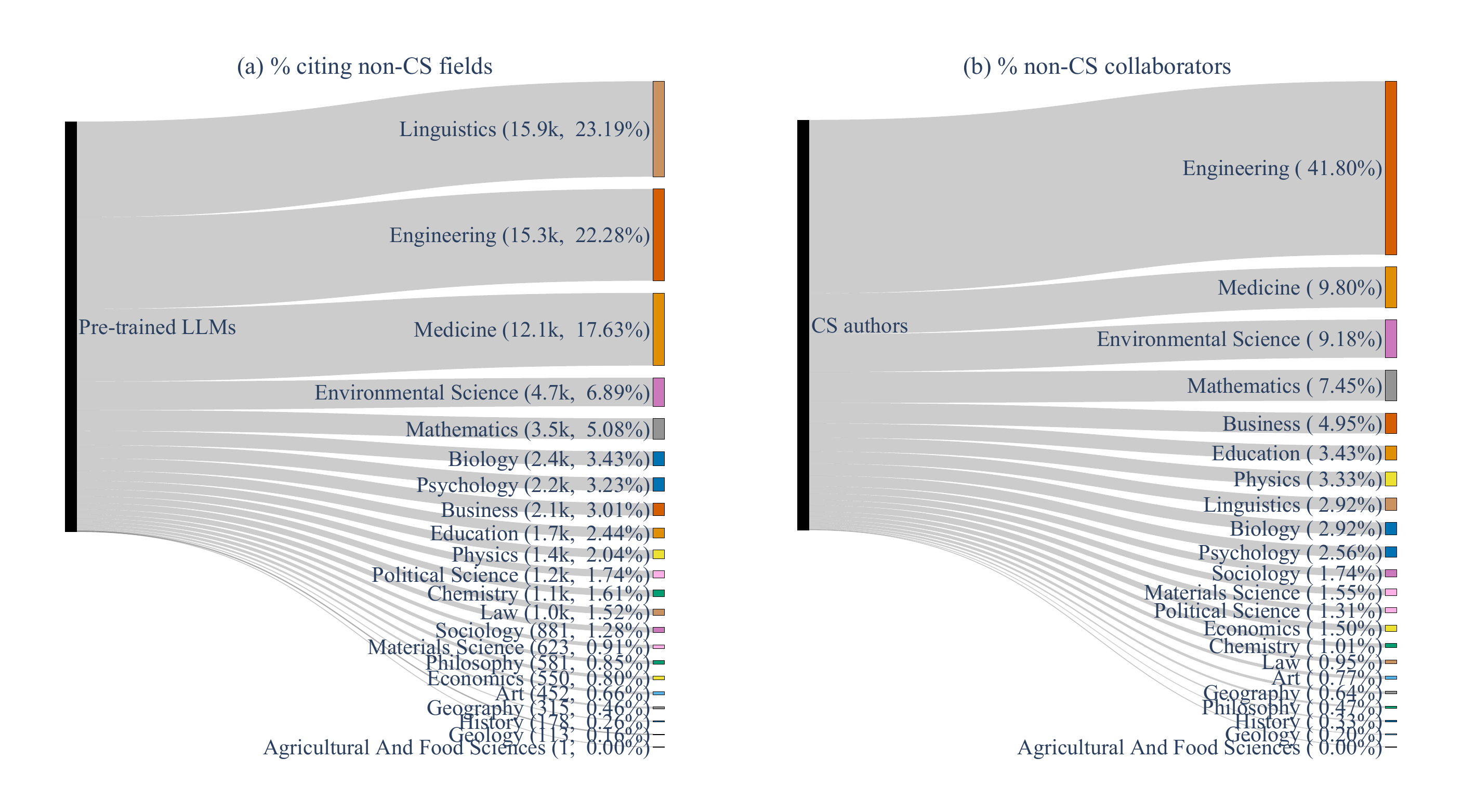}
\vspace*{-7mm}
}
    \caption{\% citations from non-CS fields to LLMs (left); \% CS authors collaborating with non-CS fields (right).}
    \label{fig:sankey_interfield_author}
\vspace*{-3mm}
\end{figure*}

\section{Data}

We initiate our study by compiling a list of $106$ LLMs 
from the Stanford Ecosystem Graphs.\footnote{\url{https://crfm.stanford.edu/ecosystem-graphs/}} This curated compilation includes the names of the LLMs selected based on their number of citations.\footnote{For details refer Appendix~\ref{app:supp_defin}} We manually identified and included in our list the papers that introduced these LLMs.\footnote{For $8$ LLMs, such as Claude or Vicuna, we cannot identify any associated papers, but we still include them in the content analysis in Section \ref{subsec:ethical_awareness}.}


\begin{table}[t]
    \centering
    {\small
    \begin{tabular}{l r}
    \toprule
    LLM Timespan & 2018 -- present \\
    \midrule
    
    \# LLMs & 106 \\
    
    \# papers citing LLMs & 148,501 \\
    
    \# citations to LLMs & 273,030 \\
    
    \bottomrule
    \end{tabular}
    }
    \caption{Statistics of LLMs and their citing papers citing.}
    \label{tab:data_stats}
\end{table}



    
    

    
    
    
    


Following 
\newcite{rungta-etal-2022-geographic}, 
we use the Semantic Scholar (S2) dataset\footnote{\url{https://www.semanticscholar.org/}} to obtain the target papers.
Specifically, we include around $148k$ papers from non-CS fields from S2 that cite the above-mentioned LLMs. Thus, 
our dataset contains structured full texts extracted from the PDFs of these papers 
along with their metadata, encompassing author details, publication year, publication venue, and field of study. In Table~\ref{tab:data_stats}, we illustrate our dataset statistics. 

We prefer S2 over other sources like the ArXiv dataset due to its broader coverage, including ArXiv papers. 
Additionally, S2 employs a field-of-study classifier (S2FOS3) based on abstracts and titles to identify the field of study (with $86\%$ accuracy). 
This dataset also contains information on paper citations, 
including the time and venue of the first publication of the citation, citation context, and citation intents (S2 dataset overview and statistics in Appendix~\ref{app:supp_defin} and Table~\ref{tab:semantic_scholar_stats}). 

Further, by analyzing author names, unique author ids, and their associations with published papers in S2, we identify the field where an author has published most frequently, which we consider as {\it their primary field of interest}.


\section{Analysis}

We use the dataset described above to examine the degree of LLM adoption in fields outside of NLP and CS, addressing three primary aspects: {\it the scope of LLM influence} (\S~\ref{subsec:impact_and_influence}), {\it their evolving utilization}(\S~\ref{subsec:utilization_patterns}), and finally, {\it their applications in these fields}(\S~\ref{subsec:ethical_awareness}). Subsequently, we offer a detailed examination of each facet.



\subsection{Breadth of LLM Adoption}
\label{subsec:impact_and_influence}

To gauge the scope of LLM influence, we address three research questions: first, we investigate {\it the extent of LLM adoption in fields beyond NLP and CS}; second, we examine {\it the depth of LLM adoption within these fields} and finally, we compare {\it LLM popularity to similar past technologies in non-CS fields}.


\label{subseq:rq1}
\noindent {\it\textbf{Q1. How widely have LLMs permeated broader academic disciplines (beyond NLP and CS)?}}
\label{subsec:llm_field_diversity}


\noindent We examine the citations in non-CS papers 
referencing LLM papers. 
If a citing paper is labeled to be in multiple fields, 
then it contributes to the citation count 
in each field. We calculate the percentage of LLM citations attributed to each non-CS field relative to the total citations from all non-CS fields. Further, we use the Gini Index~\citep{lerman1984note} on the relative citation counts to quantify the degree of deviation from a perfectly equal citation distribution across fields. A Gini index of 0 indicates a perfectly uniform distribution, while an index of 1 denotes 
all the probability mass concentrated on one value.\\[-12pt] 

\noindent{\bf Results:} Figure~\ref{fig:sankey_interfield_author}(a) is a Sankey plot of the incoming citations from non-CS fields to LLM papers $(\#citations, \%citations)$, with the width of the grey flow path representing the citation volume. 
Certain fields stand out in terms of 
LLM citations. Linguistics accounts for the maximum number of citations (23.19\%), followed by Engineering (22.28\%), Medicine (17.63\%), Environmental Science (6.89\%), and Mathematics (5.08\%). \\[-12pt]

\begin{figure}
    \centering
    \scalebox{0.85}{
    \includegraphics[width=0.5\textwidth]{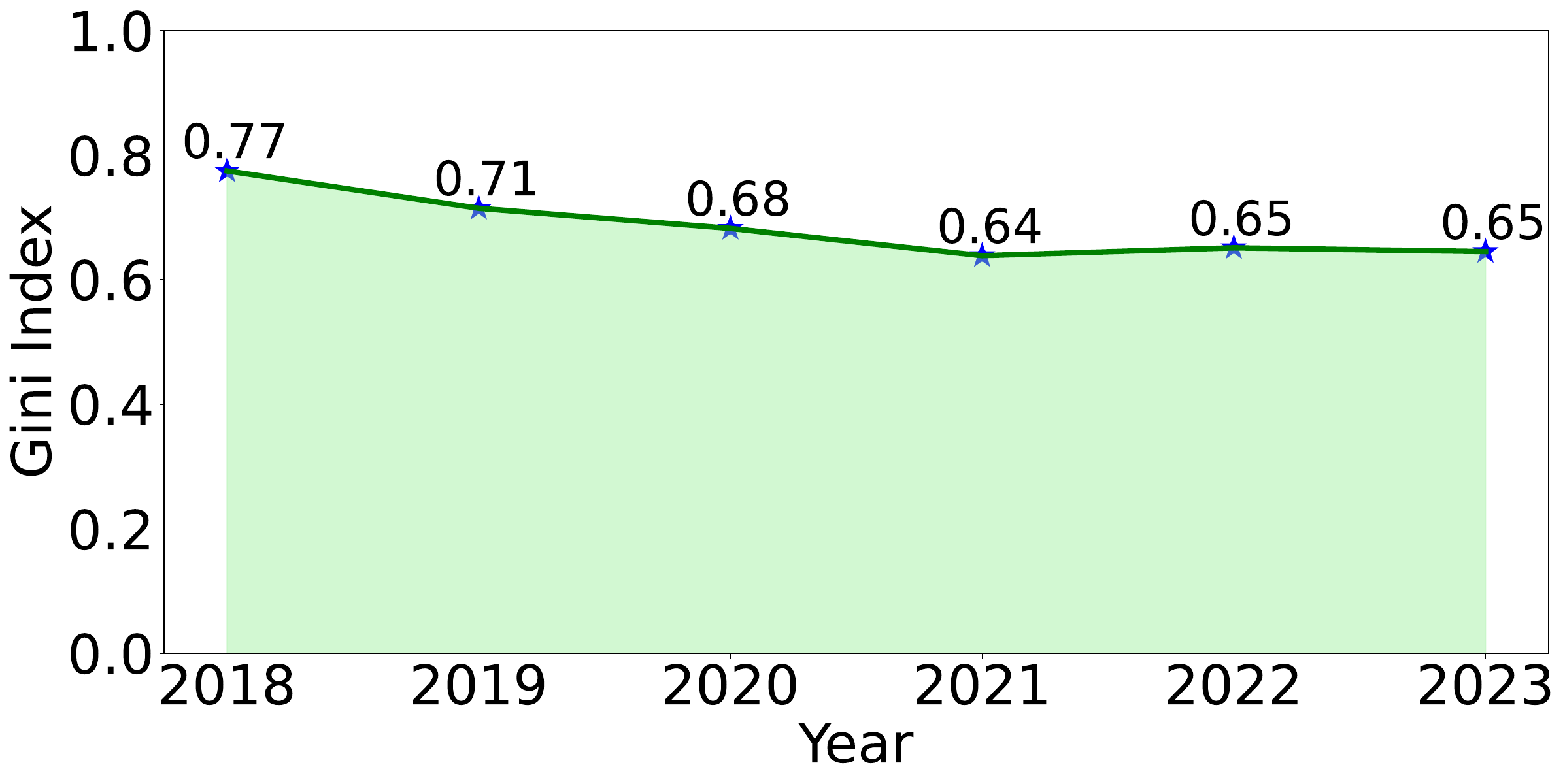}
    }
    \vspace*{-3mm}
    \caption{
    Inequality in citation distribution across fields.}
    \vspace*{-2mm}
    \label{fig:gini_diversity}
\end{figure}

\noindent{\bf Discussion:} To illustrate the evolving citation trends, in Figure~\ref{fig:temporal_incoming_llm} (Appendix~\ref{sec:appendix}), we present a year-wise analysis of citations to LLM research papers from non-CS fields. 
Our findings reveal that, in 2018, during the nascent stages of LLM research, a significant proportion of citations stemmed from Linguistics and Engineering. However, a shift becomes prominent as we progress toward 2023, with numerous non-CS fields increasingly adopting LLMs.
In Figure~\ref{fig:gini_diversity}, the Gini diversity index, declining over the years, supports this shift, indicating a broader adoption of LLMs. 

To better understand the reasons behind the citation trends of LLMs, we investigate whether high LLM citations in some fields result from interdisciplinary research. To address this aspect, in Figure~\ref{fig:sankey_interfield_author}(b), we calculate the percentage of Computer Science (CS) authors collaborating with each non-CS field relative to the total number of CS authors collaborating with all non-CS fields. Surprisingly, we find only a weak correlation (Pearson correlation: $0.28$) between the proportion of LLM citations by a field and the proportion of CS collaborators. 
This suggests that high LLM citations in some fields are likely driven more by other factors, such as 
the utility of LLMs for the exploration of research questions of interest in the field.

Further, to assess the depth of LLM penetration across each of these disciplines, examining the variations in publication volumes among these fields is important. In other words, a field with a high number of LLM citations may not be highly influenced by them when viewed in light of the total number of papers published in that field, and vice versa. This guides our exploration of the following section Q2.\\[-8pt]



\noindent {\it \textbf{Q2. To what degree have LLMs penetrated various non-CS fields?}}
\label{subsec:rq2}

\begin{figure}
    \centering
    \scalebox{0.85}{
    \includegraphics[width=0.5\textwidth]{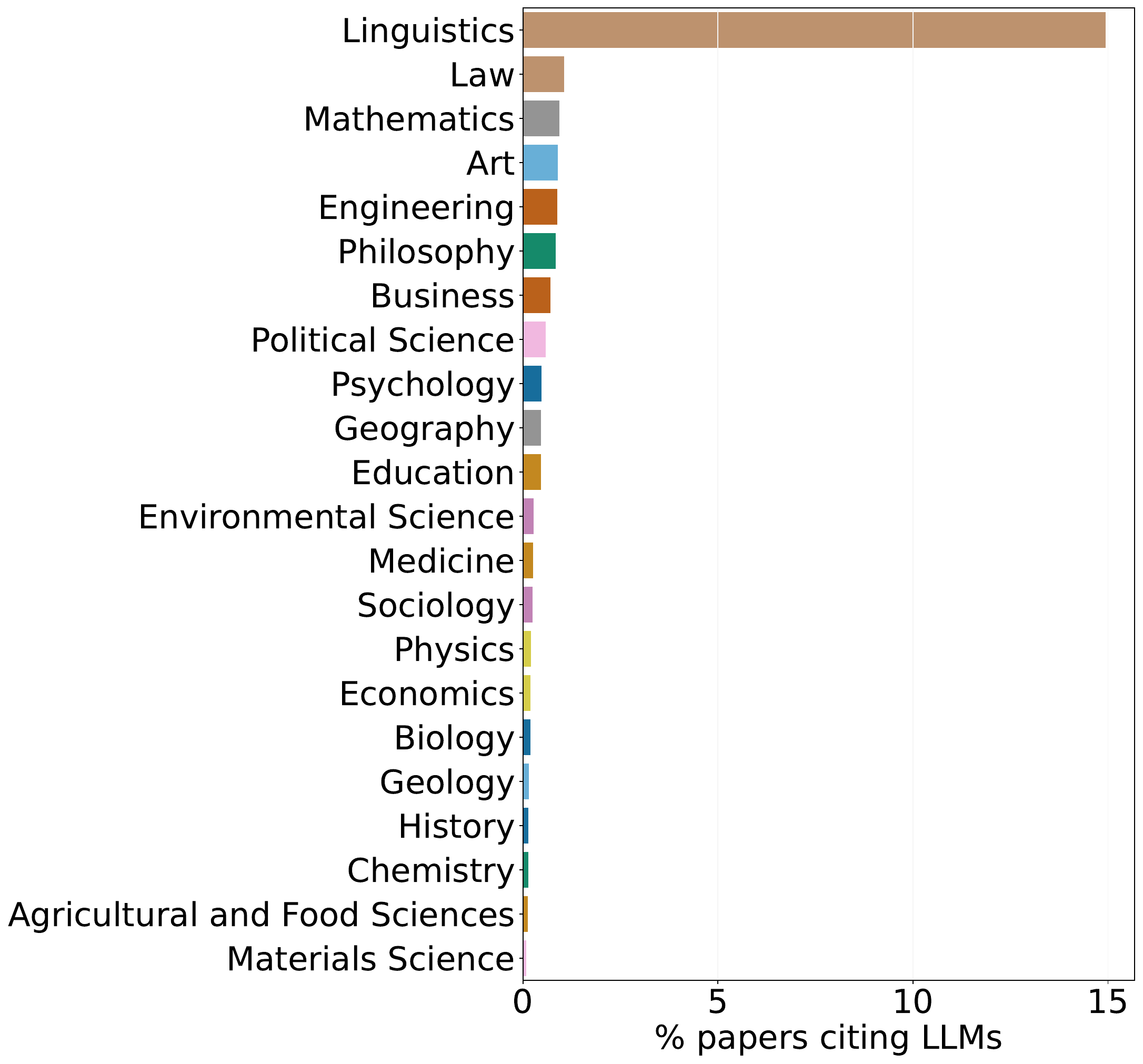}
    }
    \caption{\% Papers in non-CS fields (Y-Axis) citing LLMs.}
    \vspace*{-3mm}
    \label{fig:percent_llm_popularity}
\end{figure}

\noindent Expanding the analysis in Q1, we now investigate the popularity of LLMs within these fields. For each non-CS field, we calculate the percentage of papers citing LLMs out of the total number of papers published in that field. This acts as a rough indicator of how extensively the field is utilizing LLMs. \\[-12pt]

\noindent{\bf Results:} Figure~\ref{fig:percent_llm_popularity} shows the LLM citation counts relative to the field's publication volume.
Proportionally, Linguistics 
has the most extensive engagement with LLMs compared to other fields. It is followed by Law, Mathematics, Art, and Engineering in terms of LLM utilization. 

\begin{figure}
    \centering
    \scalebox{0.85}{
    \includegraphics[width=0.5\textwidth]{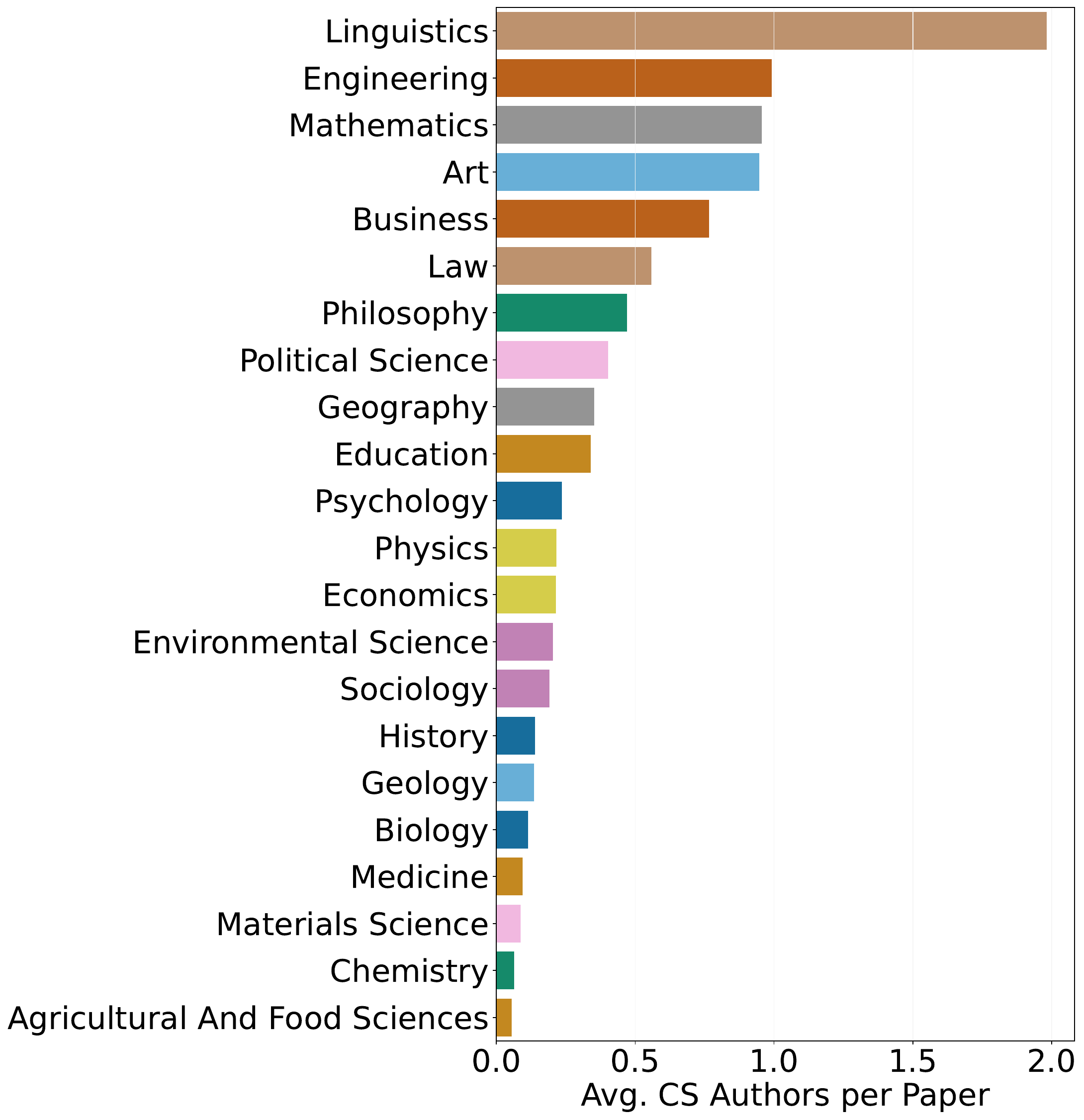}
    }
    \caption{Avg.\@ number of CS authors per paper by field (Y-Axis).}
    \label{fig:cs_authors_per_paper_per_field}
     \vspace*{-3mm}
\end{figure}

\noindent{\bf Discussion:} Previously, we noticed that while many papers in Medicine cite LLMs, their proportion (concerning the total papers in the field) is smaller than in Mathematics or Environmental Science. We further analyze the influence of CS author collaborations on research papers within each field by calculating the average number of CS authors per paper within that field since 2018.
As illustrated in Figure~\ref{fig:cs_authors_per_paper_per_field}, 
the fields with the highest average number of CS author influence are Linguistics, Engineering, and Mathematics.
We observe a positive Pearson correlation of $0.38$ between the average number of CS authors per paper in a non-CS field and the percentage of papers within that field that cite LLMs (indicating moderate correlation). This empirical evidence indicates that although the total number of CS authors collaborating within a field has little consequence; an increased number of CS authors on a research paper raises the likelihood of using LLMs in that work. \\[-8pt]

\noindent {\it \textbf{Q3. To what extent do non-CS fields embrace LLMs compared to other technologies from CS?}} \\[-10pt]
\label{subsec:rq6}

\noindent Curious as NLP researchers, we were intrigued by the reception of non-CS fields towards emerging technologies from CS beyond just LLMs. To explore this, we examine the impact of two iconic technologies: Sequence Models (RNNs and LSTMs) and Hidden Markov Models (HMMs), whose revolutionary effects in CS subfields mirror the transformative influence of LLMs today and juxtapose their wide impact with that of LLMs.  We compile all representative papers for RNNs-LSTMs and HMMs from \citet{jurafsky2000speech}, highly cited not just within NLP but also across other CS sub-fields, and count the citations they receive from non-CS fields using a methodology the same as Q1. To assess the diversity of their adoption, we use Gini indices on the citation distributions for these three technologies.



\begin{table}[t]
\centering
  \scalebox{0.85}{
  
    \begin{tabular}{l c}
    \toprule
    Cited Method & Gini Index \\
    \midrule
    
    HMMs & 0.66 \\
    
    RNN-LSTMs & 0.63 \\

    
    LLMs & {\bf 0.56} \\
    
    \bottomrule
    \end{tabular}
    }
    \vspace*{-1mm}
    \caption{Diversity (Gini Index) in citation distribution across fields for NLP Technologies.}
    \label{tab:control_diversity}
\end{table}

\noindent{\bf Results:} In Table~\ref{tab:control_diversity}, we present the diversity indices for LLMs alongside RNNs-LSTMs and HMMs. 

\begin{figure}
    \centering
    \scalebox{0.85}{
    \includegraphics[width=0.45\textwidth]{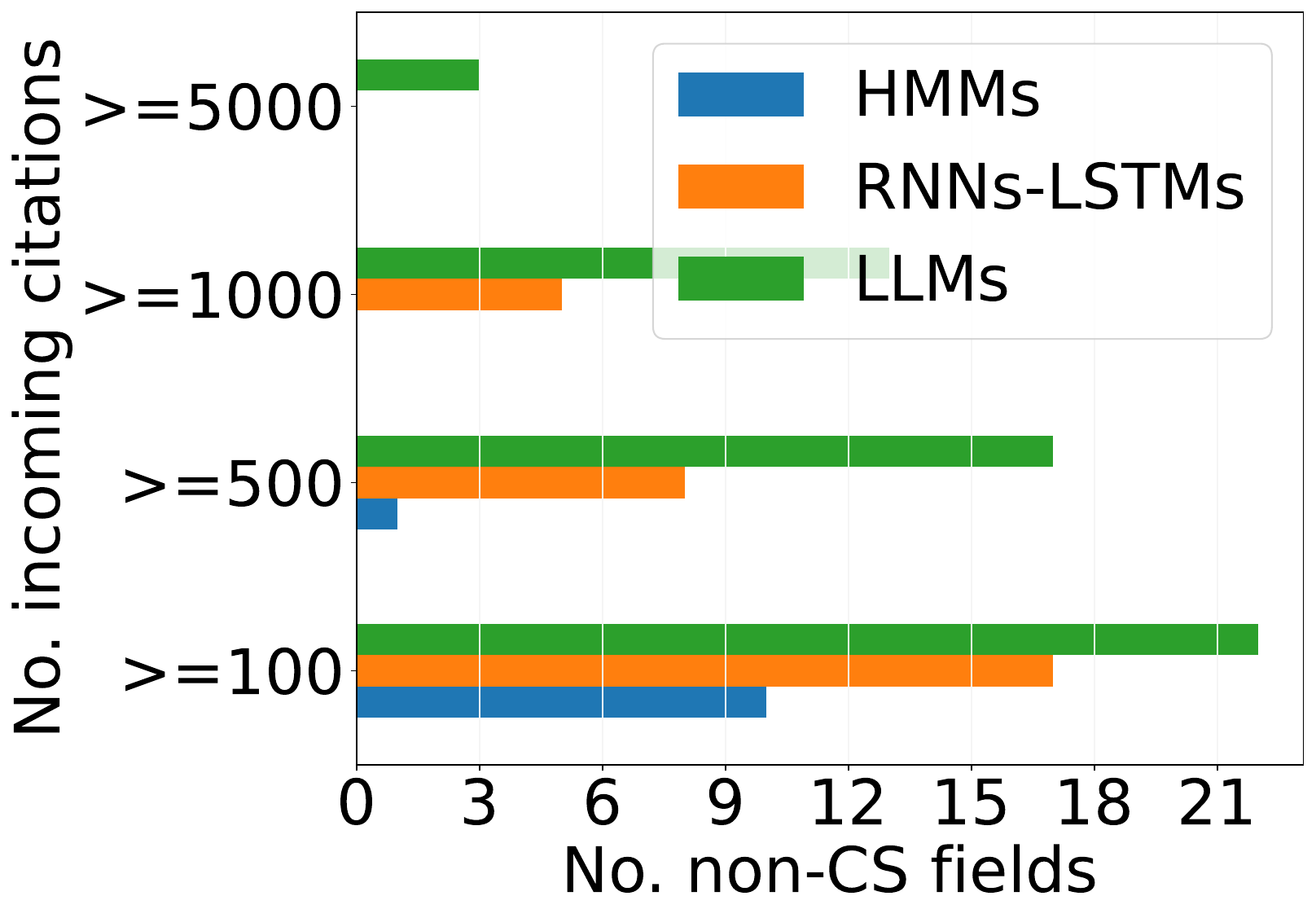}
    }
    \caption{Number of fields surpassing the threshold number of citations for different NLP Technologies.}
    \label{fig:llm_control_diversity}
    \vspace*{-2mm}
\end{figure}

\noindent{\bf Discussion:} LLMs demonstrate the lowest Gini index among these technologies, suggesting more diverse adoption across various non-CS fields. Further, to compare the popularity of LLMs with previous technologies, we set four different thresholds and show the number of fields surpassing each threshold. Figure~\ref{fig:llm_control_diversity} 
shows that LLMs receive citations more broadly across different fields, emphasizing their wider range of applications.\footnote{Refer to Appendix~\ref{app:data} for additional analysis.}


\subsection{Evolving Usage Patterns of LLMs}
\label{subsec:utilization_patterns}

To study how LLM utilization across various fields has evolved, we address two research questions: first, we identify {\it the most widely used LLMs in these fields}; second, we analyze {\it which fields have embraced newer LLMs over time}. 

\noindent {\it \textbf{Q4. What are the most popular LLMs in different non-CS fields, and why?}}\\[-12pt]
\label{subsec:rq3}

\noindent We evaluate the popularity of LLMs across 
non-CS fields by computing the annual average number of citations each LLM receives from these fields. 

\begin{figure}
    \centering
    \scalebox{0.70}{
    \includegraphics[width=0.45\textwidth]{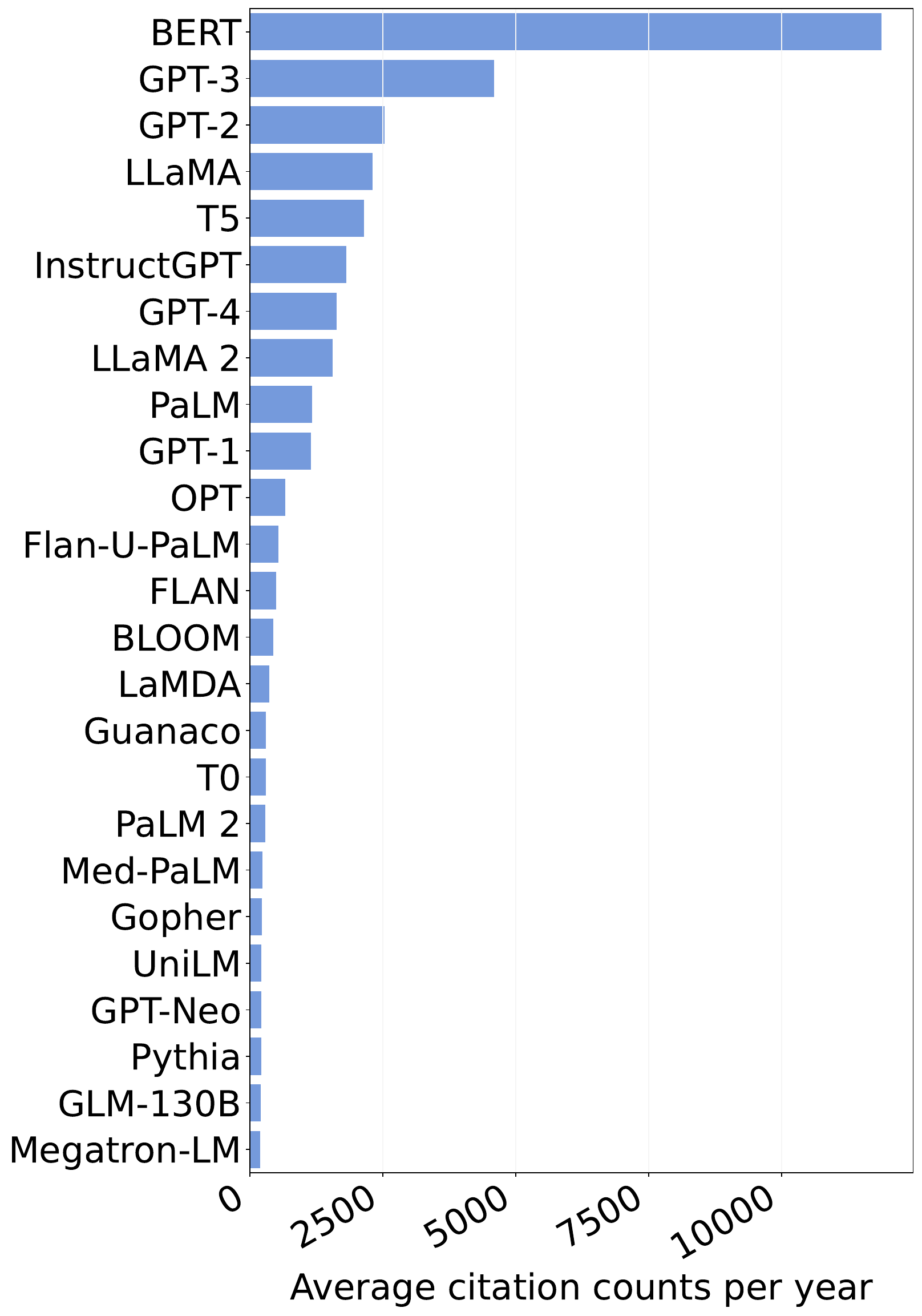}
    }
    \vspace*{-1mm}
    \caption{Popular LLMs in non-CS fields (Top 25).}
    \label{fig:popular_llms_top25}
    \vspace*{-1mm}
\end{figure}

\noindent{\bf Results:} Figure~\ref{fig:popular_llms_top25} 
shows the top 25 most popular LLMs across diverse non-CS fields~\footnote{Plot for larger list of LLMs is in Figure~\ref{fig:llm_popularity_all}, max. citation in Figure~\ref{fig:max_llm_citation} (Appendix \ref{app:data}).}.
Below, we summarize the main findings.

\noindent{\bf Discussion:} BERT, despite being published in 2019 (relatively early in the evolution of LLMs), remains the most popular LLM among non-CS fields, with the highest average citations per year. This could be because when BERT was introduced, there were fewer LLM options, leading to more citations. Consequently, many papers that incorporated LLMs during that period likely cited BERT, contributing to its higher average citation count. 
Secondly, the extensive, wide, and successful use of BERT likely made it a reliable choice for non-CS fields. In contrast, the behaviors and capabilities of newer LLMs are still undergoing exploration and evaluation.

\begin{figure}
    \centering
    \scalebox{0.85}{
    \includegraphics[width=0.45\textwidth]{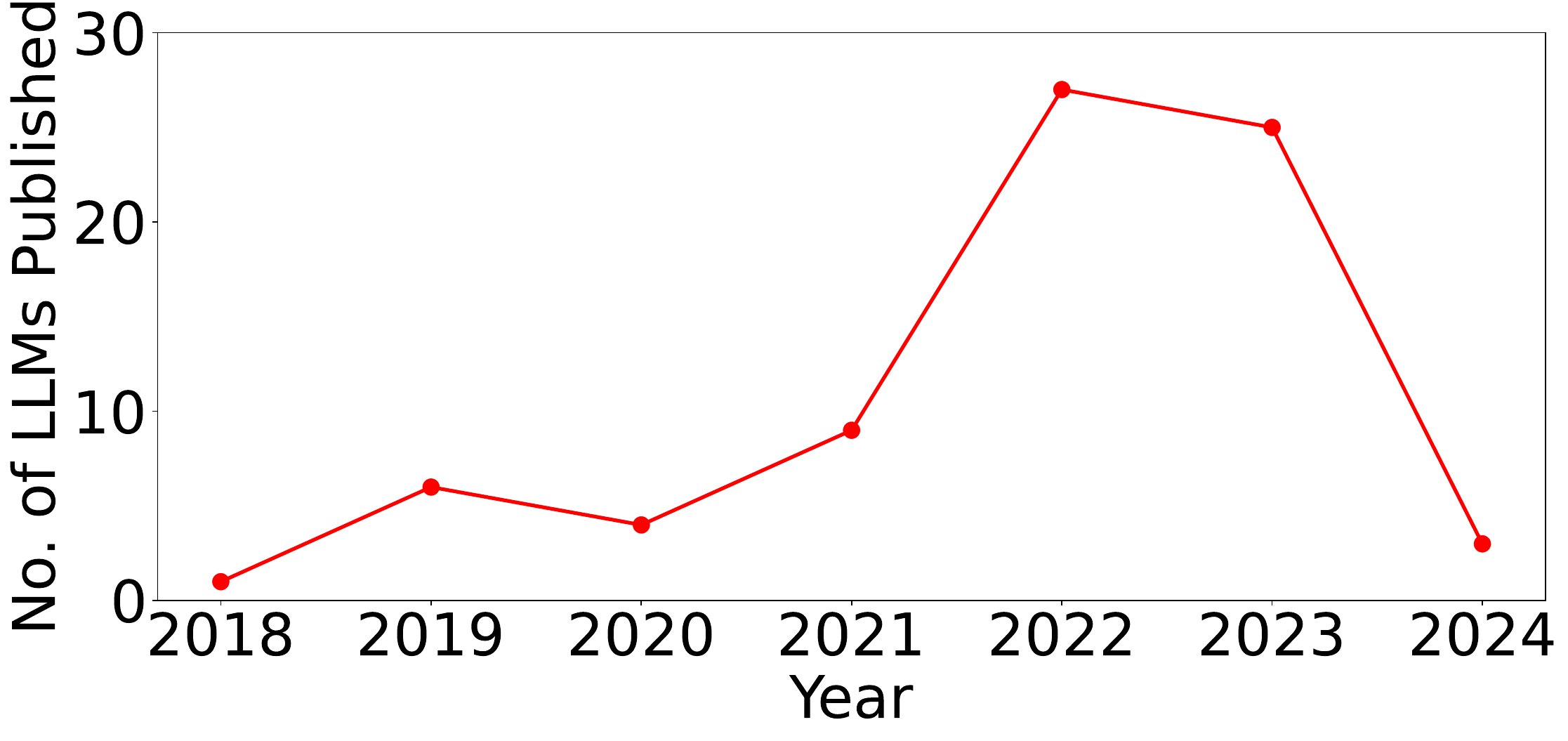}
    }
    \vspace*{-1mm}
    \caption{No. of LLMs Published over the Years (until Feb.'24).}
    \label{fig:num_llms_per_year}
    \vspace*{-2mm}
\end{figure}

In Figure~\ref{fig:num_llms_per_year}, we show the number of LLMs published in each year.
We observe that, in recent years, the proliferation of LLM research has led to the emergence of numerous new LLMs, 
resulting in some spread of citations away from the earlier favorites to 
newer LLMs (e.g., GPT-3, GPT-2, LLaMA, T5).

\noindent {\it \textbf{Q5. What is the average  LLM citation age in non-CS fields, and how does it differ across fields?}}
\label{subsec:rq4}

\noindent The LLM citation age represents the average age of the LLMs that papers in a particular field are citing. A low citation age indicates a preference for newer LLMs, while a high age suggests reliance on older ones. Fields falling in between use a mix of both.

If a paper $x_i$ from research field $c$ (i.e., $x_i \in c$) cites an LLM paper $y_j$, then the age of LLM citation (AoC) is taken to be the difference between the year of publication (YoP) of $x_i$ and $y_j$:
\[
AoC(x_i, y_j) = YoP(x_i) - YoP(y_j)
\]

We calculate the mean Age of LLM Citation ($mAoC_c$) for a field $c$ as: 
\[
mAoC_c = \frac{1}{MN}\sum_{i=1}^{M}\sum_{j=1}^{N}AoC(x_i, y_j), \forall x_i \in c
\]

\begin{figure}
    \centering
    \scalebox{0.83}{
    \includegraphics[width=0.45\textwidth]{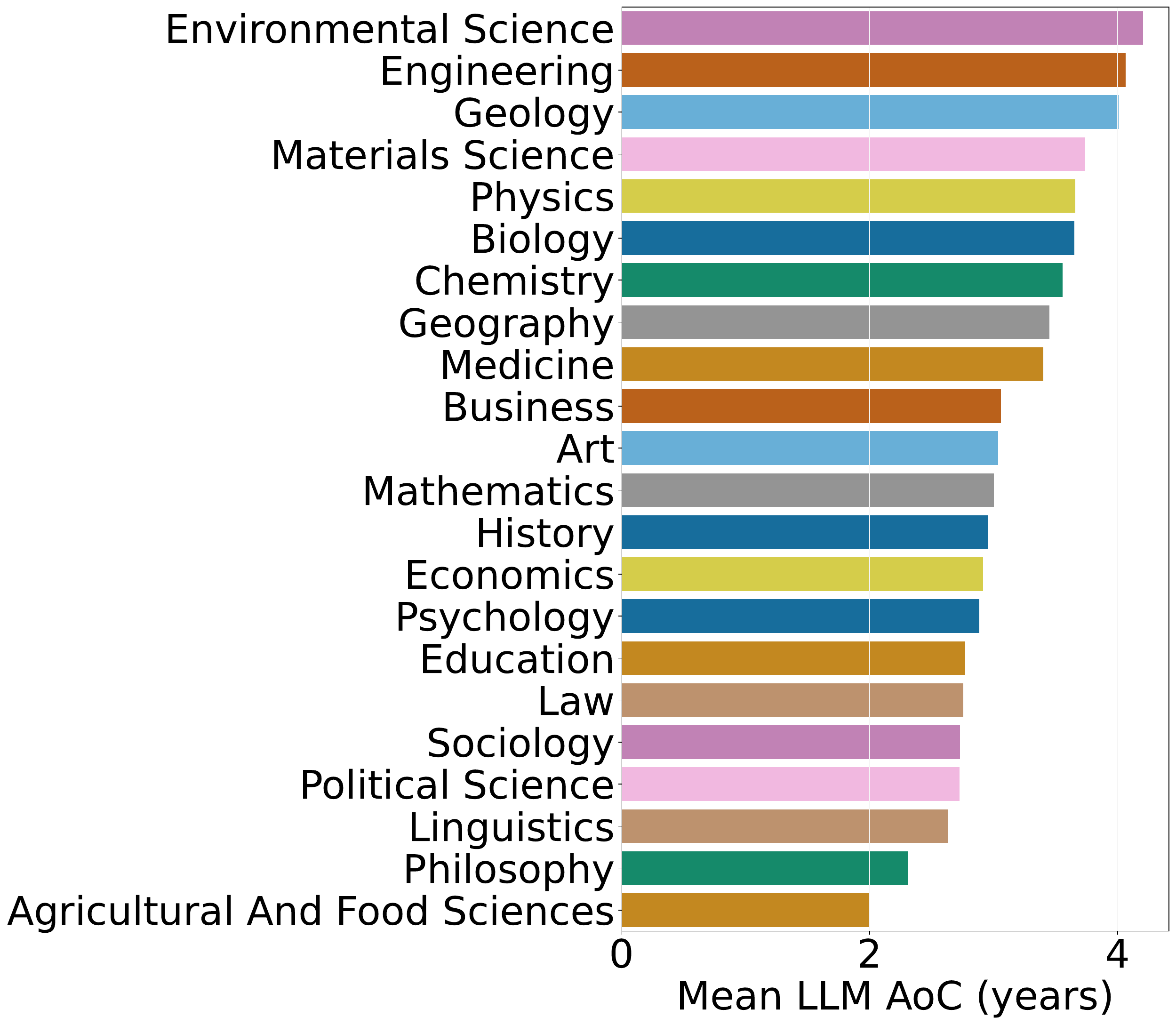}
    }
    \caption{Age-of-Citation (AoC) of 
    non-CS fields (Y-Axis).}
    \label{fig:aoc_non_cs}
    \vspace*{-2mm}
\end{figure}

\noindent{\bf Results and Discussion:} Figure~\ref{fig:aoc_non_cs} shows the Age of LLM Citation in various non-CS fields, revealing insightful perspectives on their utilization of LLMs. 

Psychology and Linguistics, having a longer history with LLMs, are now exploring newer LLMs, leading to low mAoC. Environmental Science relies mainly on older LLMs, possibly due to established performance and a lack of LLMs specifically created for the field. However, fields like Biology and Chemistry keep a balance between older LLMs while embracing newer LLMs, which are suited for their task, leading to moderate mAoC. \\[-10pt]



\subsection{Applications of LLMs in non-CS Fields}
\label{subsec:ethical_awareness}

To examine LLM applications in non-CS fields, {\it we qualitatively identify tasks where LLMs are applied}, and additionally, {\it using keyword searches, we roughly gauge the frequency with which the papers that cite LLMs in these fields mention ethical risks}. 

\noindent {\it \textbf{Q6. How are LLMs utilized in non-CS fields?}}\\[-12pt] 

\noindent Understanding these field-specific tasks, where LLMs find applications, can help the NLP community better understand, adapt, and align LLMs and research with the distinct requirements of these communities. We examine LLM usage in various non-CS fields using trigrams in paper titles, abstracts, and citation contexts. By citation contexts, we precisely mean 
sentences that explicitly mention LLMs by name or citation~\citep{anderson2023citation}. 
We further process them to identify trigrams indicative of tasks using regex-based heuristics, such as removing the stop-words and manually filtering out irrelevant trigrams. Unlike \citet{mohammad2020examining}, we use trigrams over bigrams, as trigrams offer richer contextual information beneficial for task identification within each field. 

\begin{table*}[t]
    \centering
     \scalebox{0.56}{
    \begin{tabular}{l l l l}
    \toprule
    field-of-study & \multicolumn{3}{c}{Frequent Task representative Trigrams} \\
    \midrule
    
    {\it Biology} & protein function prediction & protein structure prediction & protein sequence design \\
    
    {\it Chemistry} & compound-protein interaction prediction & molecular properties prediction & molecular dynamics simulations\\

    {\it Psychology} & mental health support & detecting rating humour & humor offense rating\\

    {\it Environmental Science} & hyperspectral image classification & image change detection & land cover classification \\

    {\it Law} & legal judgement prediction & legal case retrieval & legal case matching\\
    
    {\it Art} & visual question answering & content style text-to-drawing & design concept generation \\
    
    {\it Sociology} & whose heritage classification & daily conversational analysis & discourse relation recognition \\
    
    {\it Business} & stock price prediction & named entity recognition & graphic layout generation \\
    
    {\it Philosophy} & event causality classification & explainable causal reasoning & cheat turing test \\

    {\it Linguistics} & grapheme to phoneme & language culture internalization & zero-shot cross-lingual transfer\\
    
    {\it Mathematics} & sensing scene classification & land cover classification & stochastic differential equations \\

    {\it Physics} & low-light image enhancement & quantum machine learning & multi-speaker speech synthesis\\
    
    {\it Education} & educational question generation & automated essay scoring & generation reading comprehension \\
    
    {\it Economics} & assistive response generation & financial statement analysis & stock price prediction \\
    
    {\it Geology} & seismic data interpolation & seismic phase picking & simulation seismic waves \\
    
    {\it Engineering} & short-term load forecasting & non-intrusive load monitoring & energy consumption forecasting \\
    
    {\it Medicine} & radiology report generation & optical coherence tomography & clinical decision support \\
    
    {\it Geography} & geographic language understanding & spatially-explicit machine learning & geo-spatial knowledge graphs  \\
    
    {\it Political Science} & monitoring public discussion & processing social psychology & reframed multilingual analysis \\
    
    {\it Agriculture And Food Science} & foaming structural studies & rice yield prediction & estimating grape yield\\
    
    {\it Materials Science} & material property prediction & crystal structure generation & functional materials discovery\\
    
    {\it History} & historical event extraction & ancient latin inscription & reconstruct ancient mosaics \\
    
    \bottomrule
    \end{tabular}
    }
    \vspace*{-2mm}
    \caption{Most frequent task trigrams in LLM citing papers from non-CS fields.}
    \label{tab:llm_task}
     \vspace*{-4mm}
\end{table*}

\noindent{\bf Results and Discussion:} In Table~\ref{tab:llm_task}, we highlight the top three most frequent trigrams that serve as representative tasks in papers citing LLMs for each of the non-CS fields. This table offers a glimpse into the specific problems LLMs are employed to address across various fields. 
We observe that the fields often use LLMs to solve their field-specific problems rather than focusing on analyzing the models themselves (Refer to Appendix~\ref{app:data} for additional analysis). 
In Table~\ref{tab:llm_task_bigrams} (Appendix \ref{app:data}), we present frequent bigrams from these research papers, offering insights into the topics that interest these fields.



\begin{figure}
    \centering
    \scalebox{0.80}{
    \includegraphics[width=0.45\textwidth]{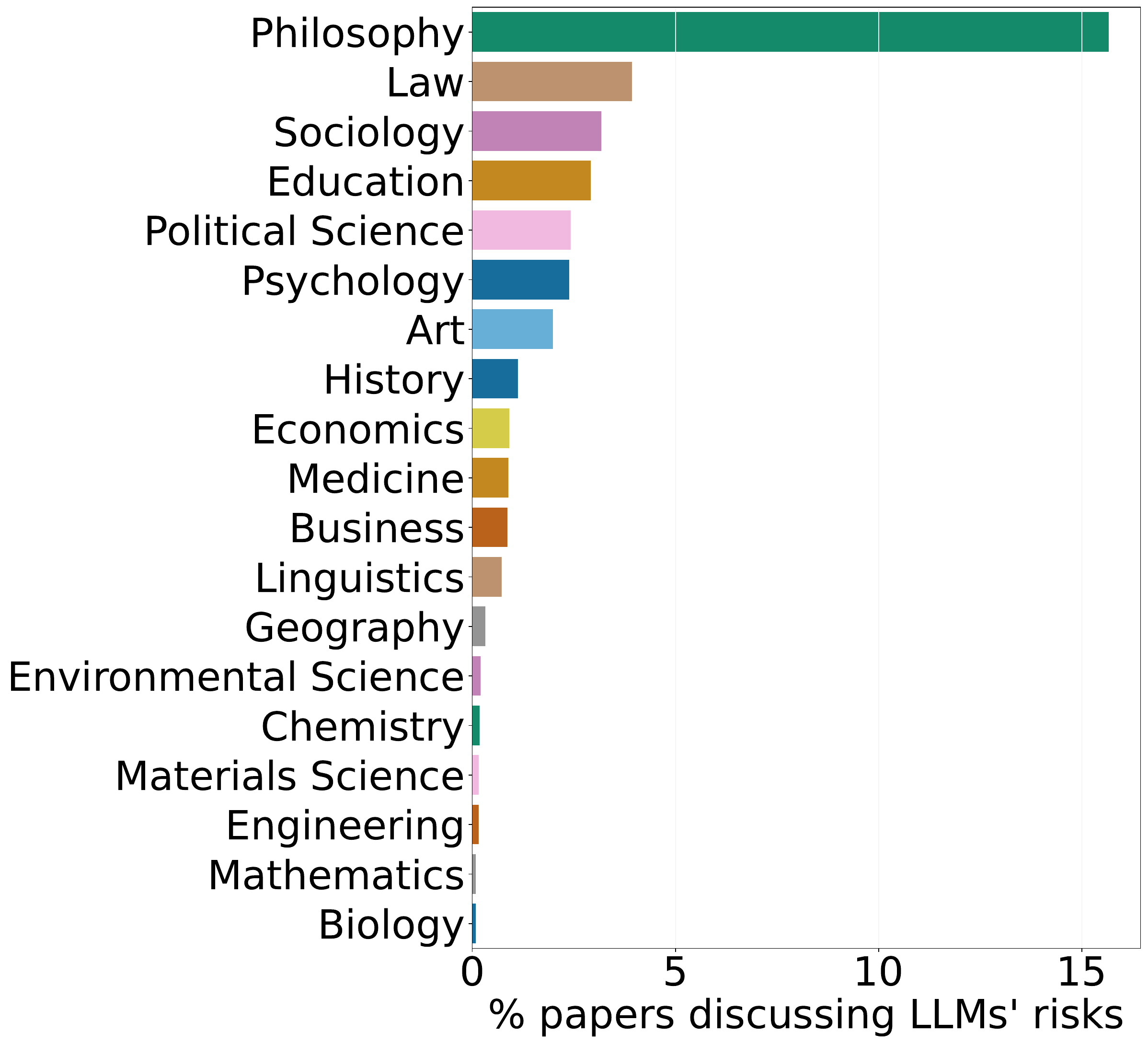}
    }
    \caption{\% of papers per field (Y-Axis) discussing LLM risks.}
    \label{fig:papers_ethical_risk}
     \vspace*{-3mm}
\end{figure}

\noindent {\it \textbf{Q7. How frequently do research papers in non-CS fields mention ethical risks?}}
\label{subsec:rq7}

\noindent Until the preceding sections, we studied the impact of LLMs on non-CS fields and their utilization trends. However, it is essential to acknowledge that LLMs come with inherent risks, such as bias and hallucination~\citep{kaddour2023challenges}. 
These concerns are particularly relevant when LLMs are applied to sensitive applications.

While we acknowledge that pinpointing research papers that address ethical concerns about LLMs from a large collection presents a significant challenge (and requires huge manual efforts), we argue that identifying the contexts in which these papers discuss risks and ethical concerns can provide a rough idea and preliminary insight into how deeply non-CS fields are engaged with the ethical implications of LLMs. Hence, examine papers that cite LLMs, aiming to identify contexts where discussions of risks and ethical considerations closely coincide with mentions of LLMs within the same sentence. Additionally, we extend our scrutiny to the titles and abstracts of these papers, assessing whether these concerns are underscored as particularly significant.

\noindent{\bf Results and Discussion:} In Figure~\ref{fig:papers_ethical_risk}, we present the percentages of papers in each field that mention the 
risks associated with LLMs in their titles, abstracts, or citation contexts at least once relative to the total number of papers citing LLMs in that field. This 
acts as an {\it approximate indicator} of the relative importance of ethical considerations within various non-CS fields. Refer to Appendix~\ref{app:data_annot} for a detailed description and manual evaluation of this analysis method.

Our analysis reveals that, on average, approximately $2.01\%$ of non-CS papers citing LLMs mentions about the ethical risks of LLMs. This finding is concerning, 
as it suggests that many non-CS fields are actively using LLMs without recognizing the ethical risks that these models entail.

\section{Discussion: Opportunities and Challenges}

In this study, we explore the adoption of LLMs across 22 non-CS fields. To analyze the large number of research papers utilizing LLMs, we employ a combination of automatic content analysis, manual scrutiny, and citation and metadata analysis. Unlike previous technologies, LLMs are finding diverse applications across disciplines such as art and history, revealing their broad scope. However, there are very few LLMs specifically tailored to meet the unique demands of these non-CS fields, highlighting a significant opportunity for improvement. Additionally, inherent risks like hallucinations associated with LLMs pose challenges. It is important to implement safe, ethical practices and communicate these risks effectively to researchers in these fields, ensuring they can make informed decisions about using LLMs. We release our artifacts under a non-commercial license \ccbyncsa.

\section*{Limitations}


In any study on field-to-field influence, having a dataset that includes both paper fields and citations is essential. This can be challenging since papers often relate to multiple fields to varying degrees, making it difficult for humans and automated systems to label them accurately. Additionally, defining distinct fields and gathering comprehensive paper sets for each field presents its own set of complexities. Nevertheless, by working with large datasets categorized based on field labels, we can derive valuable insights and identify trends in interdisciplinary research.

While citations serve as a quantifiable measure of influence \citep{siddharthan2007whose}, the degree of impact can vary between different cited works. Moreover, citation patterns are susceptible to various biases~\citep{zhu2015measuring, singh-etal-2023-forgotten}. Given the large volume of papers citing LLMs, conducting an in-depth manual examination of each paper is impractical within the scope of this study. To address this, we combine automated methods with manual verification, which offers a high-precision approximate assessment of LLM influence in non-CS fields. This methodology enables us to examine a larger set of data, providing insights into the relative influence of LLMs across diverse non-CS fields.

Lastly, the dataset used for this analysis, provided by Semantic Scholar, is extensive but not exhaustive, leaving out papers and fields beyond its scope. Additionally, Semantic Scholar relies on a classifier, along with heuristics, to categorize papers into fields of study. It is essential to recognize that this classifier has limitations and does not fully capture the nuanced degrees of association with multiple fields for a given paper. In Appendix~\ref{app:s2_reliability}, we discuss the reliability of the dataset for our analysis.

\section*{Ethics Statement}

We clarify that the number of citations used in our analysis should not be employed to diminish any specific field or its investments based on low citation counts. Decisions in science and research should rely on a multifaceted evaluation that considers aspects like popularity, relevance, resources, impact, and geographical and temporal factors. This approach prevents oversimplified interpretations and recognizes the diversity and complexity of research fields and their contributions.

\section*{Acknowledgements}

This work has been funded by the German Research Foundation (DFG) as part of the Research Training Group KRITIS No. GRK 2222. We are thankful to Hiba Arnaout and Sukannya Purkayastha for their valuable feedback on the initial draft of this manuscript.

\bibliography{custom}


\appendix


\section{Appendix: Supplementary Definitions and Discussions}
\label{app:supp_defin}

\subsection{Choosing the LLMs} 
The Computer Science community has yet to establish a universally accepted definition of LLMs~\citep{radford2019language}. The term LLM gained prominence recently amidst a paradigm shift in AI, marked by the emergence of models trained on broad data, which started with textual data but soon expanded to other data modalities (such as images, codes, proteins, etc). These models belong to a class called Foundation models~\citep{bommasani2021opportunities}.

In our analysis, an LLM refers to an impactful foundation model~\citep{henderson2023foundation} based on the transformer architecture pre-trained on massive textual datasets. We curate these models from the Stanford Ecosystem Graphs~\citep{bommasani2023ecosystem}, which contains $106$ foundation models pre-trained on textual data as of February 2024. 


\subsection{S2 Dataset Statistics and Reliability}
\label{app:s2_reliability}

The Semantic Scholar Dataset (S2) that we use in our analysis (which is the backbone of Semantic Scholar) covers all STM (Science, Technology, and Medicine) and SSH (Social Sciences and Humanities) disciplines, including biology, medicine, computer science, geography, business, history, and economics.

The February 2024 version of the S2 comprises approximately 200M papers metadata (Table~\ref{tab:semantic_scholar_stats}), sourced from diverse partners such as PubMed, Springer Nature, Taylor \& Francis, SAGE, Wiley, ACM, IEEE, arXiv, and Unpaywall etc. Originally housing 84.1M papers, the dataset is continuously updated on a monthly basis. Further, each paper in the dataset is annotated with fields of study with an annotation accuracy of 86$\%$, using a taxonomy adapted from the Microsoft Academic Graph~\citep{kinney2023semantic}.

Compared to other datasets (e.g., arXiv, Scopus, Pubmed), S2 stands out as the largest and most up-to-date open-access scholarly dataset that includes parsed text and metadata such as field of study, citation contexts, citation graph, etc. Overall, the S2 dataset is a good representative of the non-CS fields~\citep{wahle-etal-2023-cite, wahle2024citation, illia2023implementation, guo2023personalized}.

\begin{table}[t]
    \centering
    {\small
    \begin{tabular}{l r}
    \toprule    
    Timespan & 1965 -- Feb. 2024 \\
    \midrule

    \# paper metadata & 200M \\
    \# paper abstract & 100M \\
    \# author metadata & 75M \\
    \# citation instances & 2.5B \\
    \# papers full-parsed & 5M \\

    \bottomrule
    
    \end{tabular}
    
    }
    \caption{Overall Semantic Scholar Dataset Statistics.}
    \label{tab:semantic_scholar_stats}
\end{table}

\subsection{Examining LLM citation within CS}

The main objective of this work is to assess the impact of LLMs on fields outside of CS. However, in this section, we analyze a control, LLM citations within CS, to provide a broader context for our findings. To add to this, we found that the LLMs receive $\sim$$91k$ citations from within CS, which is $\sim$$5.5$ times higher than the citations it garners from Linguistics (the field that gives the highest LLM citations outside CS). This discrepancy is expected, considering the higher popularity of LLMs within CS.

\section{Appendix: Human Evaluation of LLM Citing Papers That Discuss LLM Risks}
\label{app:data_annot}

Titles and abstracts are primarily meant to provide a summary of a paper's focus, while citation contexts offer detailed insights into how the citing paper utilizes the cited work, specifically LLMs in this case. When referring to citation contexts, we mean sentences that include citation marks. More precisely, these are sentences that explicitly mention LLMs by name or citation. Hence, we use texts from these sections to filter papers that acknowledge the ethical risks of LLMs. Specifically, we look for the keywords in Table~\ref{tab:ethical_keywords} in those sections. This filtering results in 1,200 research papers in 22 fields from 148,501 papers citing LLMs. Subsequently, we manually annotate the titles and citation contexts in these 1200 papers to explore how these non-CS papers acknowledge the risks pertaining to LLMs and take steps to mitigate them. We categorize citation contexts that mention LLM-related risks into two classes: ``acknowledge'', which denotes contexts that solely acknowledge the risks, and ``mitigate'', denoting contexts that make attempts to alleviate the issues posed by LLMs (examples in Table~\ref{tab:citation_llm_risk} in Appendix~\ref{app:data}). Additionally, in Table~\ref{tab:title_ethics} (Appendix~\ref{app:data}) 
we provide a list of 10 titles from the papers that cite LLMs and 
explicitly discuss the ethical context of AI.

\paragraph{Evaluation:} To assess the reliability of the aforementioned analysis method, we conducted a manual analysis of 100 papers discussing the ethical concerns of LLMs, revealing a high recall, as only 7 papers engaged in discussions about the ethical concerns of LLMs without explicitly mentioning the LLMs. Additionally, we manually analyzed 200 papers and found no paper mentioning ethical concerns in the same context as LLMs but not discussing LLMs' ethical risks. This reveals the high reliability of our automatic method.

\begin{table}[t]
    \centering
    \begin{tabular}{l l l}
        \toprule
         ethics & risks & limitations \\
         drawbacks & bias & ethical consideration \\ 
         \bottomrule
    \end{tabular}
    \caption{Keywords to filter papers.}
    \label{tab:ethical_keywords}
\end{table}

\section{Appendix: Additional Results}
\label{sec:appendix}
\label{app:data}

\subsection{Influence of Alternate CS Method (LDA) vs. LLM beyond CS.}

In Section~\ref{subsec:rq6}, our primary focus was on models commonly used in NLP for Language Modeling, aiming to compare their influence with that of LLMs. Our motivation was to assess whether LLMs are more prevalent and influential than traditional Language Modeling methods in non-CS fields. However, in this section, we further extend our analysis to compare the influence of LLMs with that of Latent Dirichlet Allocation (LDA), a popular method originating from CS but gaining traction outside CS domains as well.

In our analysis, we identified three seminal papers on LDA~\citep{blei2003latent, blei2012probabilistic, teh2004sharing} based on their citation counts and examined the number of citations they received from non-CS fields. In total, these papers receive 8383 citations from non-CS fields cumulatively. Specifically, we found that 13 fields contribute more than 100 citations each, while 7 fields contribute over 500 citations each, and only one field contributes more than 1000 citations. Further, we find that the Gini index of the citation distribution of the aforementioned LDA papers among non-CS fields is 0.59 (higher than that of LLMs). From this analysis, we infer that LLMs are more popular in fields outside CS than LDA. 

\subsection{Usage Patterns of LLMs in non-CS Fields: Inference vs. Fine-Tuning}

We investigated the extent to which papers in non-CS fields utilize LLMs through fine-tuning or solely for inference in zero-shot settings. Specifically, we searched for the keywords "fine-tune," "zero-shot," and "inference" within the abstracts of the non-CS papers that cite LLMs. This filtering process resulted in 3675 papers. Further, we uniformly sampled and manually analyzed 50 papers out of 479 papers that mentioned fine-tuning, revealing an 86$\%$ precision. Similarly, we sampled 100 papers from the remaining 3196 papers (that did not contain the word "fine-tune" in their abstracts, and 3158 of these papers contained "zero-shot" in their abstracts) and manually analyzed them, finding a 79$\%$ precision rate in representing papers that use LLMs solely for inference.

This analysis is an approximate indicator of the use of LLMs in non-CS fields, indicating that while some of the non-CS papers indeed fine-tine LLMs on domain-specific datasets, approximately 6.6 times more papers solely use LLMs for inference purposes.

\begin{figure}
    \centering
    \scalebox{1.0}{
    \includegraphics[width=0.5\textwidth]{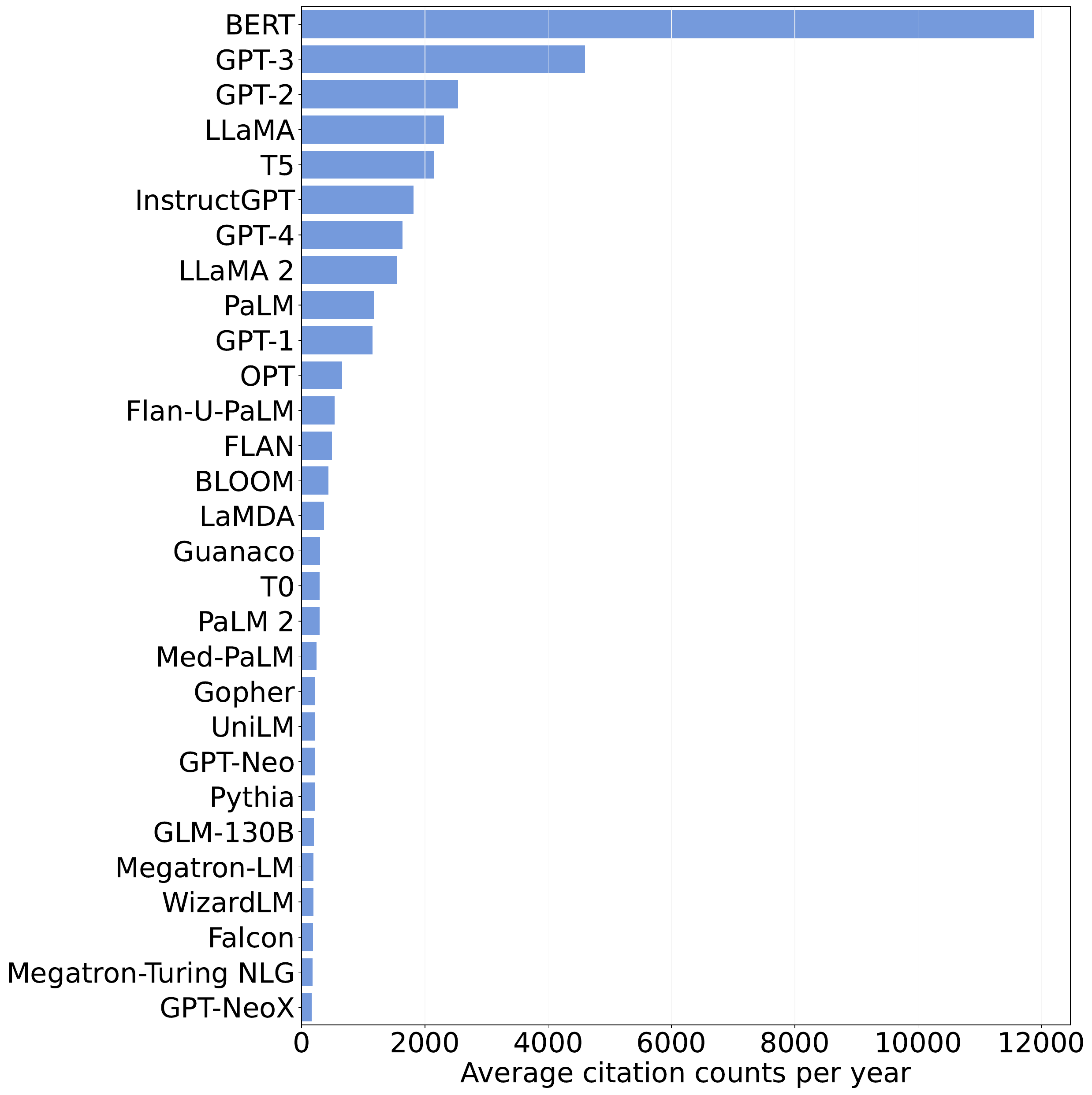}
    }
    \caption{LLM Popularity across all non-CS fields.}
    \label{fig:llm_popularity_all}
\end{figure}

\begin{figure}
    \centering
    \scalebox{0.85}{
    \includegraphics[width=0.5\textwidth]{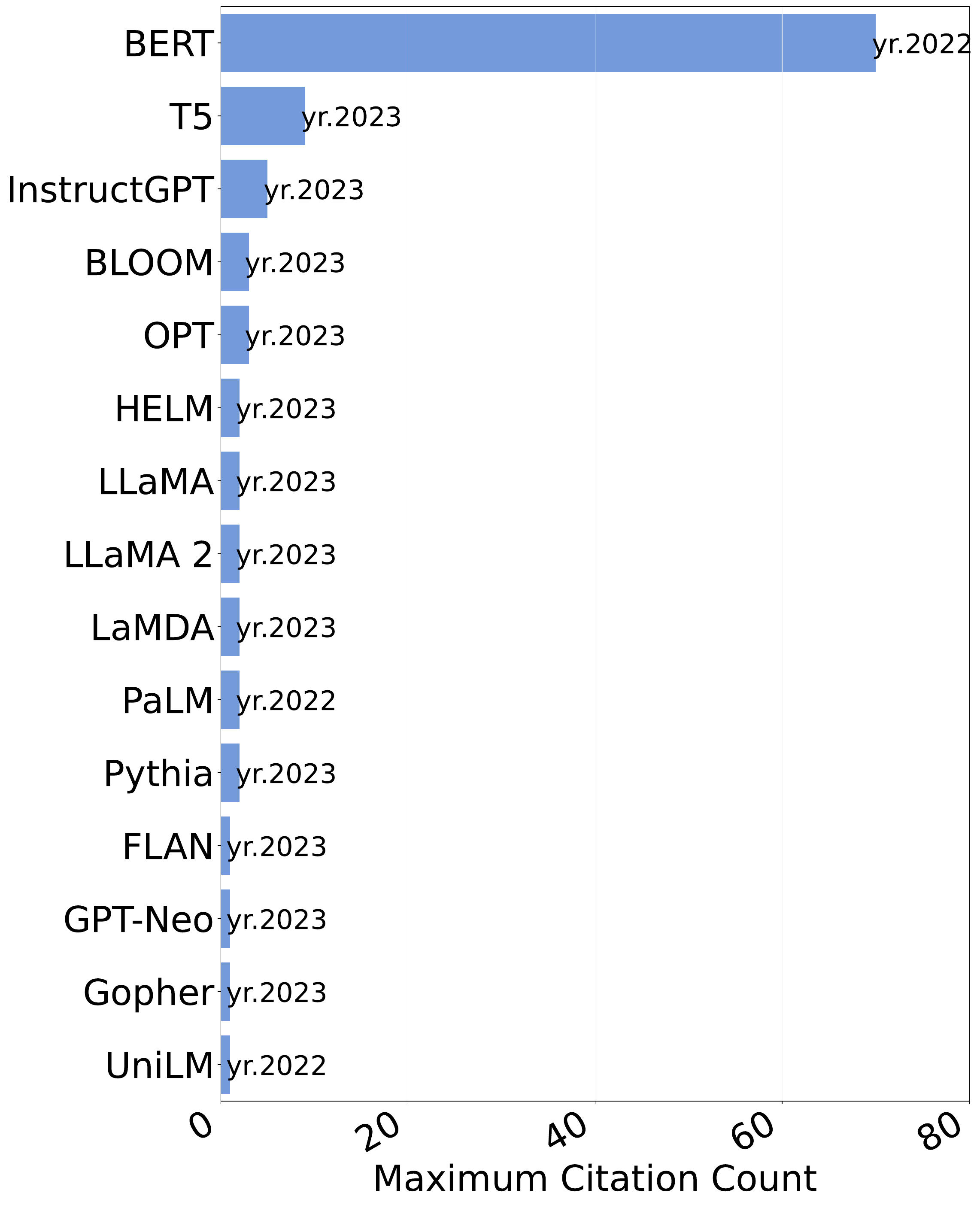}
    }
    \caption{Top 15 LLMs: Maximum Citations Across Non-CS Fields.}
    \label{fig:max_llm_citation}
\end{figure}



    
    

\begin{figure*}
    \centering
    \scalebox{0.95}{
    \includegraphics[width=1.0\textwidth]{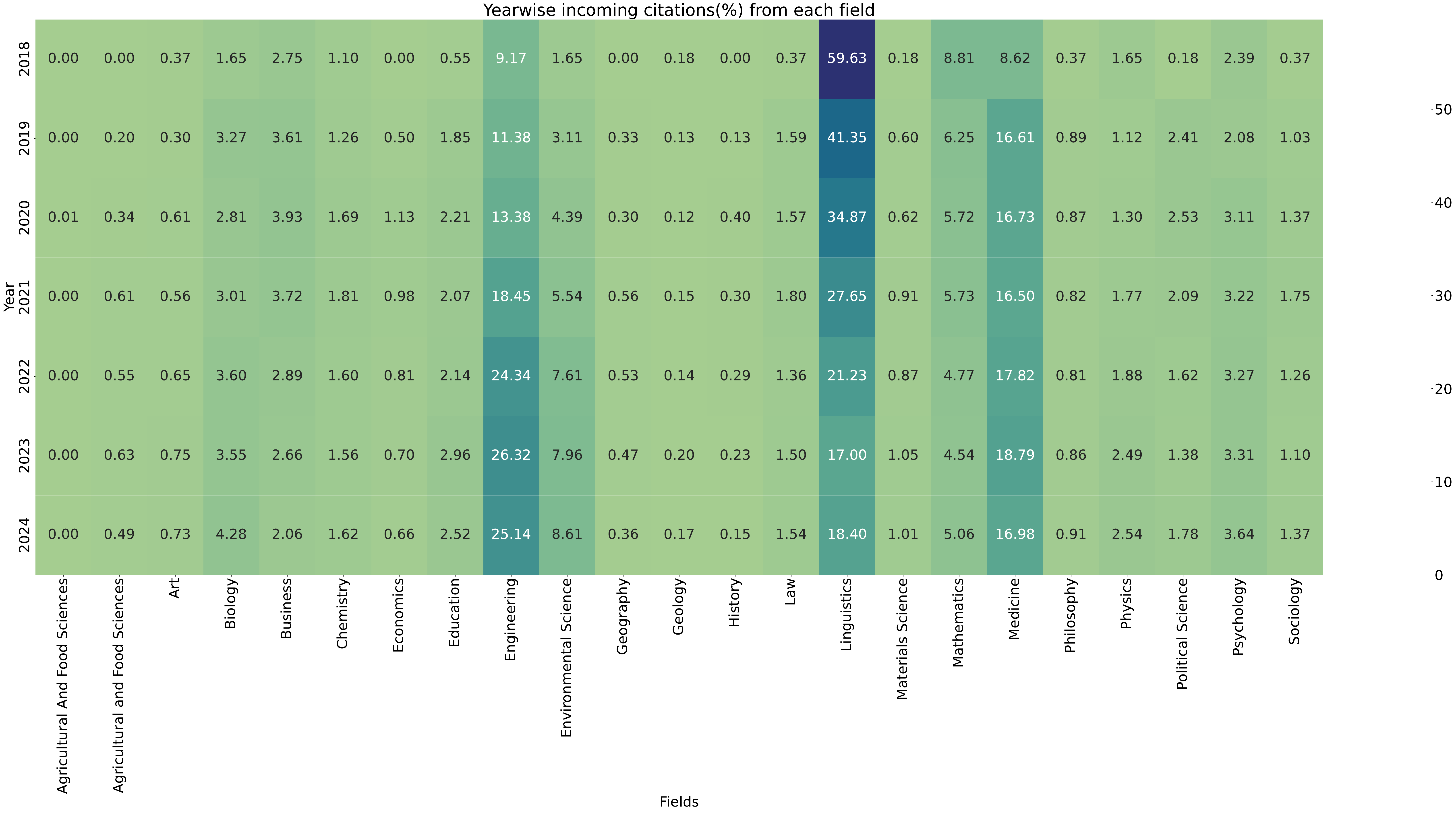}
    }
    \caption{Yearwise Incoming LLM Citations (\%).}
    \label{fig:temporal_incoming_llm}
\end{figure*}

\begin{figure*}
    \centering
    \scalebox{0.95}{
    \includegraphics[width=1.0\textwidth]{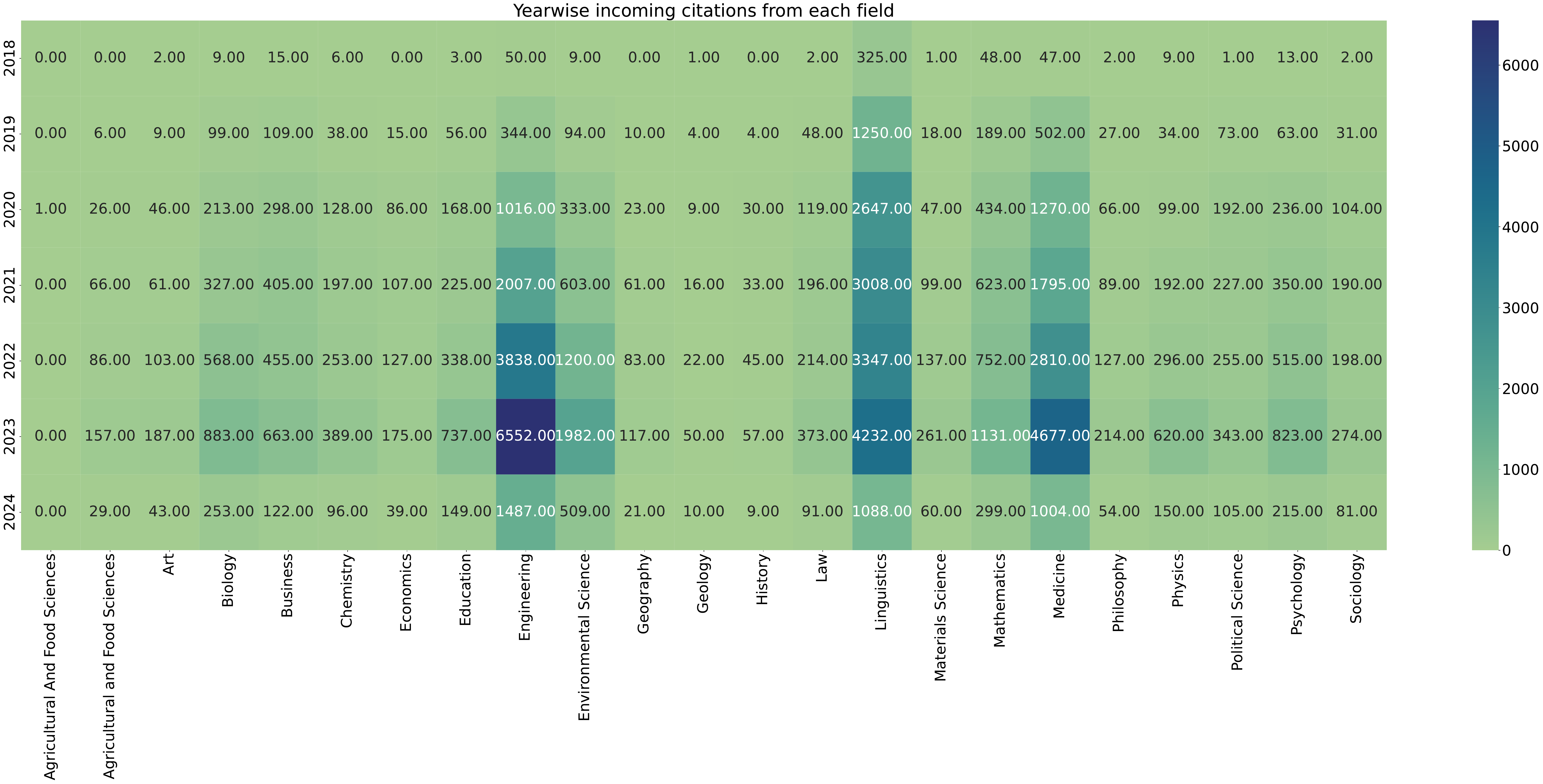}
    }
    \caption{Yearwise Number of Incoming LLM Citations.}
    \label{fig:temporal_incoming_llm_absolute}
\end{figure*}

\begin{table*}[t]
    \centering
    \small
    \scalebox{0.95}{
    \begin{tabularx}{\textwidth}{c c X X c}
    \toprule
    Field & Year & Title & Context & Class \\
    \hline
    \multirow{3}{*}{Psychology} & 2021 & An Evaluation of Generative Pre-Training Model-based Therapy Chatbot for Caregivers & However, researchers found that GPT-3 did not yield satisfying performance because it generated off-topic, confusing answers and had ethical issues, such as cultural biases in its responses. & acknowledge \\
     & 2023 & Comparing Sentence-Level Suggestions to Message-Level Suggestions in AI-Mediated Communication & Cautious voices have warned about the ethical and social risks of harm from large language models, ranging from discrimination and exclusion to misinformation and environmental and socioeconomic harms. & acknowledge \\
     & 2023 & AI Text-to-Behavior: A Study In Steerability & However, as Ray discusses, while there's a substantial promise for steerable language models, there are also crucial challenges, biases, and ethical considerations surrounding ChatGPT and similar models. & acknowledge \\

     \midrule

     \multirow{2}{*}{Medicine} & 2023 & From Military to Healthcare: Adopting and Expanding Ethical Principles for Generative Artificial Intelligence & We can adopt this principle for the ethical use of generative AI in healthcare and ensure that human involvement is maintained when more powerful generative AI systems such as ChatGPT or clinical decision support systems are in use. & mitigate \\
      & 2023 & Foundation Models in Healthcare: Opportunities, Risks \& Strategies Forward & How the use of FM-based applications may exacerbate social inequalities); and raises fundamental questions about the responsible, ethical, and safe use of such technologies going forward. & acknowledge \\

      \midrule

      Biology & 2021 & Ten future challenges for synthetic biology & The highly acclaimed natural language processing (NLP) model GPT-3 has sparked serious ethical debate due to its ability to generate highly convincing human text, even when given only a few learning points as input in a process called few-shot-learning. & acknowledge \\

      \midrule

      Sociology & 2022 & The Ghost in the Machine has an American accent: value conflict in GPT-3 & The value alignment problem is one of the more difficult areas of the field of ethical AI, but also the most critical. & acknowledge \\

      \midrule

      History & 2022 & Unified Detoxifying and Debiasing in Language Generation via Inference-time Adaptive Optimization & Moreover, such issues are found to persist across increasing model sizes, emphasizing the urgency of developing practical methods for ethical NLG. & acknowledge \\

      \midrule

      Economics & 2022 & Prismal view of ethics & Something along that line of thinking, but outside of ethical considerations, was done in machine learning for solving a considerable set of tasks with the same agent. & mitigate \\

      \midrule

      Business & 2023 & On the Planning Abilities of Large Language Models - A Critical Investigation & Of particular interest to us in this paper is the thread of efforts that aim to investigate (and showcase) reasoning abilities of LLMs, including commonsense reasoning, logical reasoning, and even ethical reasoning. & acknowledge \\
     
    \bottomrule
    \end{tabularx}
    }
    \caption{Citation contexts mentioning LLM-related ethical concerns.}
    \label{tab:citation_llm_risk}
\end{table*}

\begin{table*}[t]
    \centering
    \small
    \scalebox{0.95}{
    \begin{tabularx}{\textwidth}{l l X}
    \toprule
    Year & Field & Title \\
    \hline 

    2020 & Business & Management perspective of ethics in artificial intelligence \\

    2022 & Business & Don't ``research fast and break things'': On the ethics of Computational Social Science \\

    2022 & Philosophy & Metaethical Perspectives on `Benchmarking' AI Ethics \\

    2022 & Medicine & A scoping review of ethics considerations in clinical natural language processing \\

    2023 & Political Science & Ethics in conversation: Building an ethics assurance case for autonomous AI-enabled voice agents in healthcare \\

    2023 & Business & The ethical ambiguity of AI data enrichment: Measuring gaps in research ethics norms and practices \\

    2023 & Business & How to design an AI ethics board \\

    2023 & Business & Attention is not all you need: the complicated case of ethically using large language models in healthcare and medicine \\

    2023 & Philosophy & A method for the ethical analysis of brain-inspired AI \\

    2023 & Philosophy & A high-level overview of AI ethics \\

    \bottomrule
    \end{tabularx}
    }
    \caption{Paper titles explicitly mentioning ethical context in AI.}
    \label{tab:title_ethics}
\end{table*}


    
    
    
    



    
    

\begin{table*}[t]
    \centering
     \scalebox{0.8}{
    \begin{tabular}{l l l l}
    \toprule
    field-of-study & \multicolumn{3}{c}{Frequent bigrams representative of areas of interest} \\
    \midrule
    
    {\it Biology} & protein structure & structure prediction & structure database\\
    
    {\it Chemistry} & structural basis & molecular dynamics & structural insights\\

    {\it Psychology} & social media & mental health & emotion recognition\\

    {\it Environmental Science} & sensing images & remote sensing & object detection \\

    {\it Law} & judgement prediction & legal case & legal reasoning\\
    
    {\it Art} & stable diffusion & text-to-image generation & text-to-image synthesis \\
    
    {\it Sociology} & case study & cultural heritage & political ideology \\
    
    {\it Business} & stock price & social media & layout generation \\
    
    {\it Philosophy} & moral code & event causality & causal reasoning \\

    {\it Linguistics} & grapheme phoneme & cross lingual & social bias\\
    
    {\it Mathematics} & hyperspectral image & inverse problems & differential equations \\
    
    {\it Physics} & low-light image & quantum neural & energy physics\\
    
    {\it Education} & question generation & keyword extraction & essay scoring \\
    
    {\it Economics} & policy uncertainty & economic policy & price prediction \\
    
    {\it Geology} & seismic phase & seismic waves & phase picking \\
    
    {\it Engineering} & load forecasting & load monitoring & fault diagnosis \\
    
    {\it Medicine} & health records & radiology reports & clinical notes \\
    
    {\it Geography} & geographic language & transient chaos & geo-spatial knowledge \\
    
    {\it Political Science} & news media & public discussion & news articles \\
    
    {\it Agriculture And Food Science} & gene family & yield prediction & drought tolerance\\
    
    {\it Materials Science} & property prediction & crystal structure & feature fusion\\
    
    {\it History} & historical event  & latin inscription & ancient mosaics \\
    
    \bottomrule
    \end{tabular}
    }
    \caption{Most frequent bigrams identifying dominant areas in LLM citing papers from non-CS fields.}
    \label{tab:llm_task_bigrams}
     \vspace*{-3mm}
\end{table*}

\end{document}